\documentclass[10pt,twocolumn,letterpaper]{article}

\usepackage{cvpr}
\usepackage{times}
\usepackage{epsfig}
\usepackage{graphicx}
\usepackage{amsmath}
\usepackage{amssymb}
\usepackage{graphicx} 
\usepackage{booktabs}  
\usepackage{tabularx}
\usepackage{algorithm2e}
\usepackage{algpseudocode}
\usepackage{subcaption}
\usepackage{multirow}
\usepackage{subcaption}
\usepackage{booktabs}

\usepackage{algpseudocode}
\usepackage{color, soul, colortbl}

\definecolor{tabrow}{rgb}{0.88,0.88,0.88}
\usepackage[first=0,last=9]{lcg}
\definecolor{Gray}{gray}{0.9}
\definecolor{White}{gray}{1}

\newcommand{\vect}[1]{\boldsymbol{\mathbf{#1} }}

\newcommand{\cnn}{CNN\xspace}
\newcommand{\cnnC}{CNN \cite{gudi2015deep}\xspace}
\newcommand{\scnnit}{SCNN-IT\xspace}
\newcommand{\ocnnit}{OCNN-IT\xspace}
\newcommand{\ccnnit}{CCNN-IT\xspace}
\newcommand{\ocnn}{OCNN\xspace}
\newcommand{\ccnn}{CCNN\xspace}

\newcommand{\scnnC}{SCNN \cite{Lin_2016_CVPR}\xspace}

\newcommand{\vggC}{VGG16 \cite{Simonyan14c}\xspace}

\newcommand{\ocvprC}{OR-CNN \cite{niu2016ordinal}\xspace}
\newcommand{\rcvpr}{R-CNN\xspace}
\newcommand{\rcvprC}{R-CNN \cite{zhao2016deep}\xspace}

\usepackage[pagebackref=true,breaklinks=true,letterpaper=true,colorlinks,bookmarks=false]{hyperref}

 \cvprfinalcopy 

\def\cvprPaperID{1259} 

\ifcvprfinal\fi
\begin{document}

\title{Deep Structured Learning for Facial Action Unit Intensity Estimation}

\author
{
    \parbox{16cm}
    {
            \centering{\large Robert Walecki$^1$, Ognjen (Oggi) Rudovic$^2$,\\ Vladimir Pavlovic$^3$, Bj\"oern Schuller$^1$  and Maja Pantic$^{1,4}$}\\
        {
            \normalsize$^1$ Department of Computing, Imperial College London, UK\\
            $^2$ MIT Media Lab, Cambridge, USA \\
            $^3$ Department of Computer Science, Rutgers University, USA\\
            $^4$ EEMCS, University of Twente, The Netherlands           
        }
    }
}


\maketitle

\begin{abstract}

We consider the task of automated estimation of facial expression intensity. This involves estimation of multiple output variables (facial action units --- AUs) that are structurally dependent. Their structure arises from statistically induced co-occurrence patterns of AU intensity levels. Modeling this structure is critical for improving the estimation performance; however, this performance is bounded by the quality of the input features extracted from face images. The goal of this paper is to model these structures and estimate complex feature representations simultaneously by combining conditional random field (CRF) encoded AU dependencies with deep learning. To this end, we propose a novel Copula CNN deep learning approach for modeling multivariate ordinal variables. Our model accounts for $ordinal$ structure in output variables and their $non$-$linear$ dependencies via copula functions modeled as cliques of a CRF. These are jointly optimized with deep CNN feature encoding layers using a newly introduced balanced batch iterative training algorithm. We demonstrate the effectiveness of our approach on the task of AU intensity estimation on two benchmark datasets. We show that joint learning of the deep features and the target output structure results in significant performance gains compared to existing deep structured models for analysis of facial expressions.
\end{abstract}

\section{Introduction}
\label{sec:intro}
Automated analysis of human facial expressions aims to make inference about affective states, emotion expressions, pain levels, etc., from face images of the target person. 
Facial expressions are typically described in terms of configuration and intensity of facial muscle actions using the Facial Action Coding System (FACS) \cite{ekman2002facial}. FACS defines a unique set of 30+ atomic non-overlapping facial muscle actions named Action Units (AUs) \cite{kohn2009cvpr}, with rules for scoring their intensity on a six-point ordinal scale. Using FACS, nearly any anatomically possible facial expression can be described as a combination of AUs and their intensities. 

Recent advances in deep neural networks (DNN), and, in particular, convolutional models (CNNs) \cite{gudi2015deep}, have allowed to completely remove or highly reduce the dependence on physics-based models and/or other pre-processing techniques, by enabling the ``end-to-end'' learning in the pipeline directly from input images. While the effectiveness of these models has been demonstrated on many general vision problems \cite{krizhevsky2012imagenet,szegedy2013deep,sun2014deep}, only baseline tasks such as expression recognition and AU detection \cite{liu2013aware,zhao2016deep,khorrami2015deep} and AU intensity estimation \cite{gudi2015deep} have been investigated. All of them, however, follow the traditional ``blind deep learning'' paradigm that relies on large labeled training datasets (e.g., 100K+ samples in \cite{parkhi2015deep}).  Yet, in the facial data domain obtaining accurate and comprehensive labels is typically prohibitive. For instance, it takes more than an hour for an expert annotator to code AUs' intensity for 1 sec of face video.  Even then, the annotations are highly biased and have low inter-annotator agreement. Coupled with large variability in imaging conditions, facial morphology, dynamics of expressions, this has resulted in the lack of suitable large datasets for effective deep model learning.

To improve deep learning for facial expression analysis and, in particular, intensity estimation of facial AUs, from available (annotated) facial images, we exploit and combine two modeling approaches: structured learning and data-sharing (e.g., between multiple datasets). We focus on the AU intensity estimation as the intensities are very difficult to annotate manually (high number of AUs and their intensity levels) but are of critical importance for high level interpretation of facial expressions. This, inevitably, entails a scarcity in available annotated data. Furthermore, the AU intensities are highly  imbalanced due to the highest intensity levels occurring rarely and varying considerably among subjects. Finally, the dynamics of AUs also vary across contexts (e.g., in facial expressions of pain vs expressions of basic emotions). 

To tackle these challenges, we first constrain the deep CNN models by imposing their structure at different levels. Specifically, we model the network output (i.e., different AUs) jointly as ordinal variables to account for the monotonicity constraints in the (discrete) intensity levels of each AU. 
Also, explicitly modeling the relations between co-occurrences of AU intensity levels has been addressed for binary outputs only (e.g., for object detection \cite{Lin_2016_CVPR}), and not for multi-level intensities. In this work, we  model the AU intensity relations by allowing them to be $non$-$linearly$ related -- in contrast to present models that account only for linear dependencies. We do so by means of copula functions \cite{braeken2007copula}, known for their ability to capture highly non-linear dependencies through a simple parametrization. The notion of the copula functions has previously been explored for modeling of structured output \cite{walecki2016copula} but not in the context of structured deep learning. 

To efficiently model these two types of structure within our deep CNN model, we borrow the modeling approach of conditional graph models (Conditional Random Fields -- CRFs) to define ({\bf ordinal}) unary and ({\bf copula}) binary cliques in the output graph (i.e., the output layer of the deep net), which are then learned jointly with the CNN layers. Note that several approaches to combining CNNs and CRFs have been proposed \cite{shen2015shadow,zheng2015conditional,dai2015boxsup}. However, these model a different type of (spatial) dependencies, and, more importantly, deal only with (object) detection tasks- thus can not be directly scaled to the multi-class ordinal classification problems, as addressed here. Our main contributions can be summarized as follows:

\begin{itemize}
    \item We propose a novel structured deep CNN-CRF model for joint learning of multiple ordinal outputs. The data structure is seamlessly embedded in the deep CNN via an output graph, capturing the ordinal structure in AU intensity levels via ordinal unary cliques, and non-linear dependencies between the network outputs via the copula binary cliques. We show that this model learns better the target AUs from scarce and highly imbalanced data compared to existing deep models.
    \hspace{-0.3cm}
    \item Joint learning of the deep CNN and target dependency structure (CRF) in our model is challenging and can easily lead to overfitting if standard learning is applied. To ameliorate this, we propose a novel approximate training: balanced-batch iterative training that carefully feeds the model with balanced variety of subjects, AU intensity levels and their co-occurrences during learning. We show that this is critical for the model's performance and leads to efficient learning.
    \hspace{-0.3cm}
    \item To leverage annotations from multiple datasets efficiently, our approach augments the learning of the shared marginals (AUs) across multiple datasets. This, in turn, results in models that are more robust to imbalanced and scarce data. 
\end{itemize}
We show on benchmark datasets of naturalistic facial expressions, coded in terms of AU intensity, that our approach outperforms by a large margin related deep models applicable to the target task. 

\begin{figure*}[t!]
	\centering
	\includegraphics[width=0.7\linewidth]{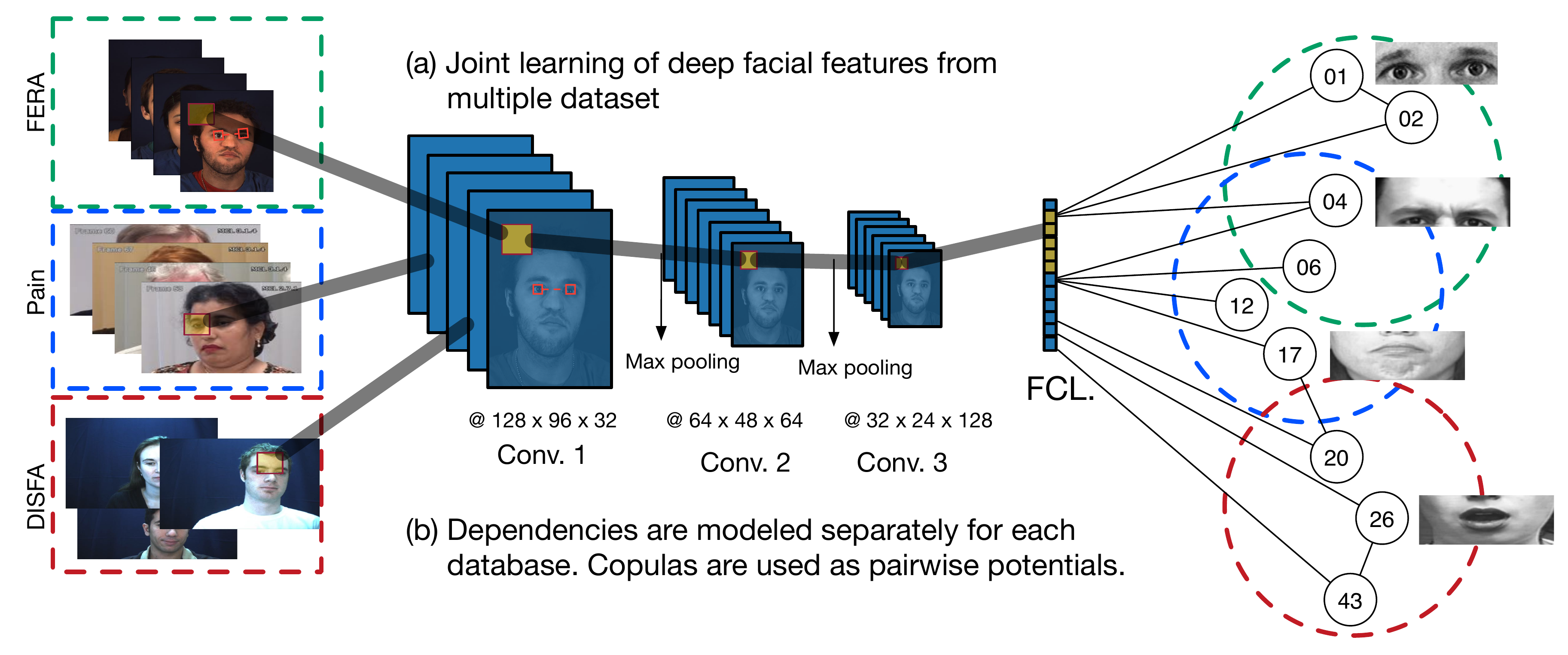}
    \caption{The proposed model pipeline. The input is a preprocessed face image, and the outputs are the model likelihood values for each intensity level of each AU. The CNN features are jointly learned for estimation of intensities of all AUs and the parameters of the unary potentials are shared. The pairwise potentials, however, model the AU dependencies that are specific to the context of the database. 
        }
    \label{fig:intro}
\end{figure*}

\section{Related Work}
\subsection{Facial Action Unit Intensity Estimation}
Estimation of AU intensity is often posed as a multi-class problem approached using  Neural Networks \cite{kapoor2005multimodal}, Adaboost \cite{bartlett2006automatic}, SVMs \cite{lucey2007investigating} and belief network classifeirs  \cite{liu2014facial}. Yet, these methods are limited to a single output, thus, a separate classifier is learned for each AU -- ignoring the AU dependencies. This has been addressed using the multi-output learning approaches. For example, \cite{nicolle2015facial} proposed a multi-task learning for AU detection where a metric with shared properties among multiple AUs was learned. Similarly, \cite{sandbach2013markov} proposed a MRF-tree-like model for joint intensity estimation of AUs. This method performs a two step learning -- by first obtaining the intensity scores for each AU independently, followed by the MRF-graph optimization -- aimed at capturing the AU relations. The proposed  Latent-Trees (LTs) \cite{kaltwang2015} for joint AU-intensity estimation capture higher-order dependencies among the input features and multiple target AU intensities. More recently, \cite{walecki2016copula} proposed a multi-output Copula Ordinal Regression approach for estimation of AU intensity, where the co-occurring AU intensity levels are modeled using the statistical framework of copula functions. However, these methods are highly dependent on the feature pre-processing, involving (dense) facial point tracking and extraction of hand-crafted image features. More importantly, these cannot deal with high-dimensional input features. To this end, this paper investigates alternative approaches based on CNNs for the target task.

\subsection{CNN Models for Facial Expression Analysis}
So far, only a few works addressed the task of facial expression recognition using CNNs. \cite{liu2013aware} introduced an AU-aware receptive field layer in a deep network, designed to search subsets of the over-complete representation, each of which aims at best simulating the combination of AUs. Its output is then passed through additional layers aimed at the expression classification, showing a large improvement over the traditional hand-crafted image features such as LBPs, SIFT and Gabors. Another example is \cite{gudi2015deep}, where a CNN is jointly trained for detection and intensity estimation of multiple AUs. The authors proposed a network architecture composed of 3 convolutional and 1 max-pooling layers. More recently, \cite{zhao2016deep} introduced an intermediate region layer that is able to learn region specific weights of CNNs. The region layer returns an importance map for each input image and the network is trained for joint AU detection. All these methods focus solely on feature extraction while the network output remains unstructured. By contrast, we capture the output structure by means of a CRF graph that explicitly accounts for ordinal and non-linear relations between multiple outputs. Note that ordinal modeling has been attempted in the context of deep networks for the age estimation task \cite{niu2016ordinal}. However, in contrast to our model, this method does not handle multi-output multi-class problems.

\subsection{Structured Deep Models}

Structured models allow us to learn task-specific constraints and relations
between output variables directly from the data (see \cite{nowozin2011structured}). Recently, this has been a focus of research within deep learning -- an attempt to regulate the network output. This is typically achieved by combining multi-output CNNs and graph models such as CRFs and MRFs. For instance, DeepLab-CRF \cite{shen2015shadow} combines a CNN and fully connected CRF, where the binary cliques are used to model relations between image color and location. More recently, \cite{Lin_2016_CVPR} proposed a fully connected CRF with $linear$ binary cliques to capture semantic correlations between neighboring image patches, showing its effectiveness on the image segmentation task. Other applications of structured CNNs include image restoration \cite{eigen2013restoring}, image super-resolution \cite{dong2014learning}, depth-estimation \cite{eigen2015predicting}, and image-tagging \cite{chen2015learning}. However, to the best of our knowledge, the deep structred learning has not been attempted before in context of facial expression analysis, and in particular, intensity estimation. Also, while the structured models mentioned above may be applicable to the target task, the key difference to our \ccnn model is that they fail to model ordinal structure in their CRF model - which is critical when dealing with ordinal variables. Also, since these methods deal with binary outputs, they assume linear relations in the binary cliques of a CRF. This can easily be violated when dealing with multi-class outputs, as in our case. To this end, we propose non-linear dependence modeling using the framework of copula functions.

\section{Structured Deep CRFs: Methodology}

Fig.\ref{fig:intro} summarizes our deep structured learning approach. We assume here several settings. In the first setting, given an input face image, we first apply a pre-defined CNN network layer to the (normalized) input image, in order to generate a feature map. The learned deep features are of a lower resolution than the original image because of the down-sampling operations in the pooling layers. To embed the target structure, we place a CRF graph on the (fully) connected output layer of our network. Here, each output (AU) of the network represents a node in this graph, and relations between different nodes (AUs) are modeled using pairwise connections in this CRF. To leverage information from multiple datasets, we propose a data-augmented learning approach (the second setting). In this approach, the CNN layers are trained using data from multiple datasets simultaneously, resulting in enriched feature representation. As these datasets may contain non-overlapping sets of AUs, the model output will be a union of all these AUs, thus,  instead of having multiple ``weak'' models, we arrive at a single shared model for multiple AUs. However, it is important to mention that for each combination of AUs (dataset-specific), we learn different dependencies in CRF pairwise connections, as their dynamics may vary considerably across the datasets. On the other hand, modeling of the marginals/nodes in the graphs is performed jointly, by sharing the model parameters of the overlapping AUs in these datasets.

For simplicity, we start with the notation that describes a single dataset as $\mathcal{D} = \lbrace \vect{Y}, \vect{X}\rbrace$ (we extend this to multiple datasets in Sec.\ref{sec:multi_data}). $\vect{Y} = [\vect{y}_1, \ldots, \vect{y}_i, \ldots, \vect{y}_N]^T$ is comprised of $N$ instances of multivariate outputs stored in $\vect{y}_i = \lbrace \vect{y}^{1}_i, \ldots \vect{y}^{q}_i, \ldots \vect{y}^{Q}_i\rbrace$, where $Q$ is the number of AUs, and $\vect{y}^{q}_i$ takes one of $\{1,...,{L^q}\}$ discrete intensity levels of the $q$-th unary potential. Furthermore, $\vect{X} = [\vect{x}_1,\ldots,\vect{x}_i,\ldots,\vect{x}_N]^T$ are input images that correspond to the combinations of labels in $\vect{Y}$. 

\textbf{Deep Facial Features:} In our experiments, we first use a CNN to extract the feature map $f_d(\vect{x}, W)$ from an input image $\vect{x}$, where the network parameters are defined by $W$. We used 3 convolutional layers containing 32, 64 and 128 filters. The filter size was set to $9\times9$ pixel followed by ReLu (Rectified Linear Unit) activation functions. We also used max pooling layers with a filter size of $2\times2$ after each convolutinonal layer. The last component of the CNN is the fully connected layer (fcl) which returns 128 facial features. These parameters have been found via a validation procedure (Sec.\ref{sec:exp}).

\textbf{Structured CRFs:}
\label{sec:crf}
We assume a graph with unary and binary cliques in our CRF\cite{lafferty2001conditional}. Specifically, we introduce a random field with an associated graph ${\mathcal G}= (V,{\mathcal C})$, where nodes $v \in V, \vert V \vert = Q$, correspond to individual AUs and cliques $c \in {\mathcal C}$ correspond to subsets of dependent AUs modeled using the copula functions. 
The conditional likelihood for image $\vect{x}$ having the labels $y$ is then defined as follows:
\begin{equation}
        \label{eq:ll}
        P(\vect{y}|\vect{x},\Omega)=\frac{1}{Z(\vect{x})}\exp{\big[-E(\vect{y},\vect{x},\Omega)\big].}
\end{equation}
Here, $Z(\vect{x})=\sum_{\vect{y}^*}\exp{\big[-E(\vect{y}^*,\vect{x},\Omega)\big]}$ is the partition function and the energy function is defined by a set of unary and pairwise potential functions.
\begin{equation}
        \label{eq:energy}
        \small
        E(\vect{y},\vect{x},\Omega)=  \sum\limits_{q \in V} {U(\vect{y}^q,f_d,\phi^q)}  + \sum\limits_{(r,s) \in E}^{} {V(\vect{y}^r,\vect{y}^s,f_d,\theta^{r,s}).}
\end{equation}
where $U$ is the unary potential function and $V$ the pairwise potential function. The parameters of $U$ and $V$ are $\phi$ and $\theta$, respectively. The input features are computed using $f_d(\vect{x},W)$ where $x$ is the input and $W$ the weights of the network.

\subsection{Unary potentials}
\label{or}
To impose increasing monotonicity constraints on the AU intensity levels, we formulate the unary potentials using the notion of ordinal regression \cite{Agresti1994Ordinal}. Let $l\in\{1,\dots,L\}$ be the ordinal label for the intensity level of the $q$-th AU. We employ the standard threshold model:
\begin{equation}
        \vect{y}_*^q = {\beta ^q}{f_d(x,W)}^T + {\varepsilon ^q},{\vect{y}^q} = l{\mkern 1mu} {\mkern 1mu} {\rm{\,\,iff\,\,}}\psi _{l-1}^q < \vect{y}_*^q \le \psi _{l}^q.
\end{equation}
where $\beta^q$ is the ordinal projection vector, ${\psi^q_l}$ is the lower bound threshold for count level $l$ (${\psi^q_0}=  - \infty < {\psi^q_1} < {\psi^q_2}... < {\psi^q_{L-1}} < {\psi^q_{L}}=+ \infty$). By assuming that the error (noise) terms $\varepsilon^q$ are Gaussian with zero mean and variance $(\sigma^q)^2$, their normal cumulative density function (cdf) is $F(z^{q}) = Pr(\varepsilon{^q<}z^{q}) = \int_{ - \infty }^{z^{q}} {{\cal N}(\xi;0,1)d\xi }$, and the probability of AU $q$ having intensity $l$ is defined as:
\begin{equation}
        {\rm{ Pr}}(\vect{y}^q = l|f_d(x,W),\phi^q) = F(z^{q}_l) - F(z^{q}_{l - 1}).
\label{eq:hor}
\end{equation}
where $z_k^q = \frac{{(\psi _k^q - {\beta ^q}{{f_d(x,W)}^T})}}{{{\sigma ^q} }}$. The model parameters are stored in $\phi^q = \{{\psi^q_1},{\psi^q_2},\dots,{\psi^q_{L-1},
\beta^q,\sigma^q}\}$. Finally, the unary node potentials in our structured deep CRF are defined as:
\begin{equation}
        {U(\vect{y}^q,{\vect{x},W,\phi^q})} = {\rm{ Pr}}(\vect{y}^q = l|f_d(x,W),\phi^q).
\label{eq:una}
\end{equation}
Note that these ordinal potentials embed the label structure in our graph -- this is in contrast to existing structured deep CRFs \cite{shen2015shadow,eigen2013restoring,dong2014learning,chen2015learning}, which typically use the soft-max/sigmoid function. 

\subsection{Pairwise potentials}
\label{cor}
The structured deep CRFs reviewed in Sec.\ref{sec:crf} focus on modeling of binary co-occurrence patterns, and the use of linear binary potentials. Yet, in case of multiple intensity levels, various and highly non-linear co-occurrence patterns are expected (e.g., for two AUs, there are $6\times6$ possible configurations). To this end, we propose a more powerfull modeling of these dependencies using the copula functions\cite{shih1995inferences}. 

The main idea of copulas is closely related to that of histogram equalization: for a random variable $y^q$ with (continuous) cdf $F$, the random variable $u^q := F(y^q)$\footnote{Sometimes we omit dependence on $f_d(x|W)$ for notational simplicity.} is uniformly distributed on the interval $[0, 1]$. Using this property, the marginals can be separated from the dependency structure in a multivariate distribution \cite{NIPS2008_COP}. In the context of structured learning, the copula functions allow us (i) to easily model non-linear dependencies among the outputs, and (ii) do so independently of their marginal models. The latter is particularly important when designing efficient learning algorithms for deep learning (see Sec.\ref{sec:learning}).

Formally, a copula $C({u^1},{u^2}, \ldots {u^Q})$: ${{\rm{[0, 1]}}^Q} \to {\rm{[0, 1]}}$ is a multivariate distribution function on the unit cube with uniform marginals \cite{opac-b1100913}. When the random variables are discrete, as is the case with the AU intensity levels, we can construct the joint distribution for discrete variables as:       
\begin{equation}
\label{pdfcop}
\begin{array}{l}
\Pr ({y^1} = {l^1}, \ldots ,{y^Q} = {l^Q}) = \\
\Pr ({\psi _{{l^1} - 1}} < y_*^1 < {\psi _{{l^1} }}, \ldots ,{\psi _{{l^Q} - 1}} < y_*^Q < {\psi _{{l^Q}} })\\
 = \sum\limits_{{c_1} = 0}^1 { \ldots \sum\limits_{{c_Q} = 0}^1 {{{( - 1)}^{{c_1} +  \ldots  + {c_Q} } }} F(z_{{l^1} - {c_1}}^1,...,z_{{l^Q} - {c_Q}}^Q)} \\
 = \sum\limits_{{c_1} = 0}^1 { \ldots \sum\limits_{{c_Q} = 0}^1 {{{( - 1)}^{{c_1} +  \ldots  + {c_Q} } }} {C_\theta }(u_{{l^1} - {c_1}}^1,...,u_{{l^Q} - {c_Q}}^Q)}.
\end{array}
\end{equation}
where $u_{{l^q} - {c_q}}^q = F(z_{{l^q} - {c_q}}^q), c_q \in \{0,1\},$ is defined in Sec.\ref{or}, and $\theta$ are the copula parameters, as defined below. It is important to note that the the joint density model induced by the copula is conditioned on the deep features $f_d(x|W)$, i.e., $ F( {y^1}, \ldots ,{y^Q} ) \leftarrow F( {y^1}, \ldots ,{y^Q} \vert f_d(x|W) )$. This, in contrast to the models in \cite{sandbach2013markov,li2013unified} that rely solely on the AU labels, allows the deep features to directly influence the dependence structure of AUs, and the other way round, during learning. Under this formulation, for the binary case, the model reduces to:
\begin{equation}
\begin{array}{c}
\Pr ({y^r} = {l^r},{y^s} = {l^s}) = F(z_{{l_r}}^r,z_{{l_s}}^s)\\
 + F(z_{{l_r} - 1}^r,z_{{l_s} - 1}^s) - F(z_{{l_r} - 1}^r,z_{{l_s}}^s) - F(z_{{l_r}}^r,z_{{l_s} - 1}^s).
\end{array}
\label{eq:pairco}
\end{equation}
We use these joint probabilities to define binary cliques in our CRF model as:
\begin{equation}
\begin{array}{c}
{V(\vect{y}^r,\vect{y}^s,\vect{x},W,\theta^{r,s})}=\Pr ({y^1} = {l^1},{y^2} = {l^2}|f_d(x|W),\theta^{r,s}).
\end{array}
\label{eq:pairco}
\end{equation}

One specific benefit of copulas is that they can model different forms of (non-linear) dependency using simple parametric models for $C(\cdot)$. We limit our consideration to the commonly used Frank copula~\cite{genest1987frank} from the class of Archimedean copulas, defined as:
\begin{equation}
        \small
        {C_\theta }({u^r},{u^s}) =  
        - \frac{1}{\theta }
            \ln 
            \left( {
                    1 + \frac
                    {\left(\exp{(-\theta u^r})-1\right) \left(\exp{(-\theta u^s})-1\right)}
                    {\exp{(-\theta})-1}
            } \right).
\end{equation}
The dependence parameter $\theta  \in ( - \infty , + \infty )\backslash \{ 0\}$, and the perfect positive/negative dependence is obtained if $\theta  \to  \pm \infty$. When $\theta  \to 0$, we recover the ordinal model in Eq.\ref{eq:hor}.

\subsection{Learning and Inference}
\label{sec:learning}
\begin{figure}[ht!]
	\centering
	\includegraphics[width=0.8\linewidth]{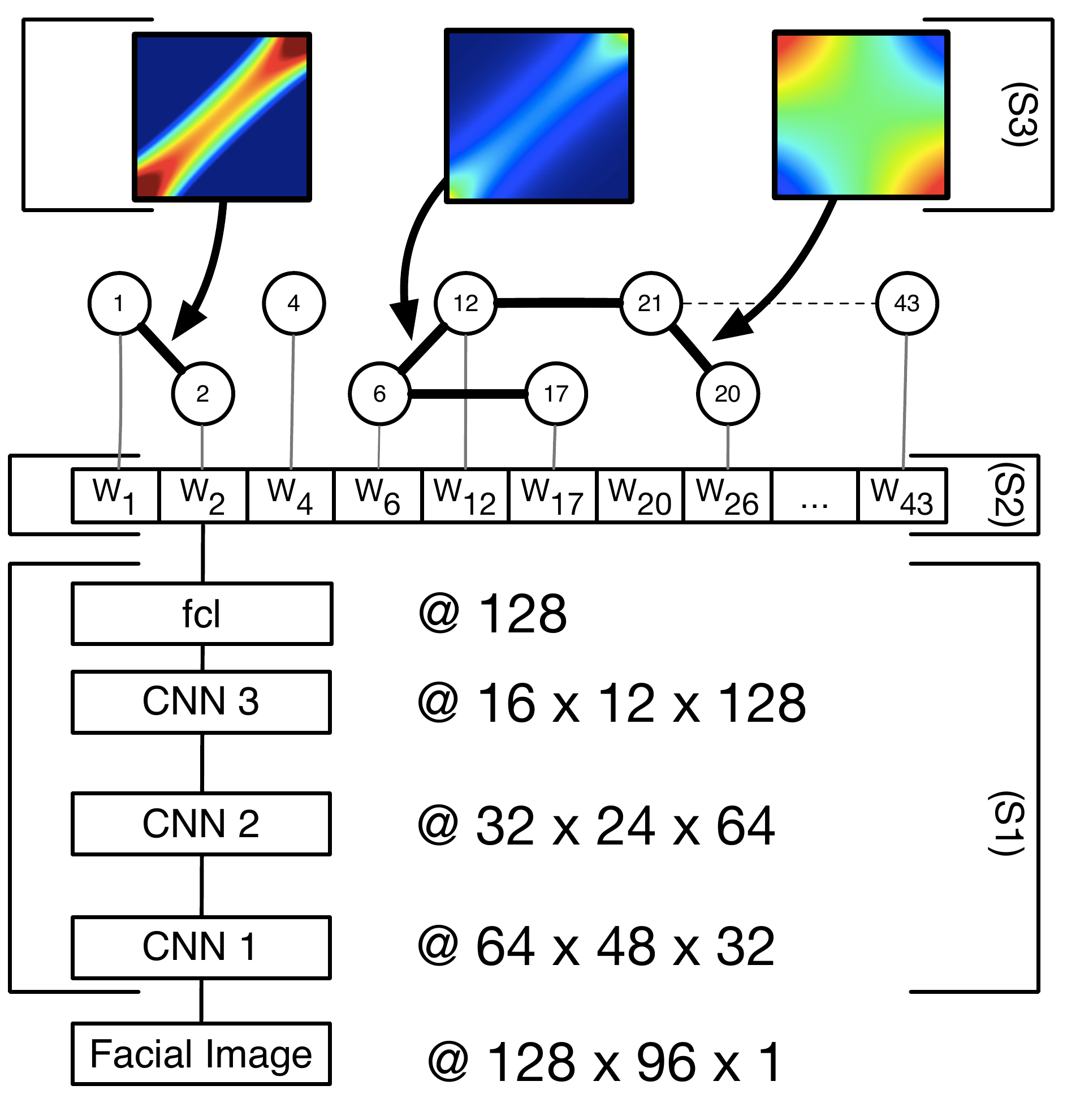}
	\caption{Three step parameter learning of the \ccnn model. The input to the network is a facial image and the global negative log likelihood is optimized in an iteratively manner. First, the wights of the CNN are optimized in step 1 (S1). In the second step (S2), we optimize the parameter of the unary potentials and finally, in step 3 (S3), we optimize the parameter of the pairwise potentials. The frank copula density function is shown for a strongly correlated pair of AUs (1\&2), for a weakly correlated pair of AUs (6\&12) and for a negatively correlated pair (21\&20).}
    \label{fig:learning}
\end{figure}
Optimizing the network parameters can be done a in naive way by minimizing the (regularized) negative log-likelihood of Eq. (\ref{eq:ll}). However, this is prohibitively expensive as it involves computation of the normalization constant $Z$, which, in case of 10 AUs, would involve $2^{10}$ evaluations of the copula functions. We mitigate this by resorting to the approximate methods based on piece-wise training of CRFs \cite{Lin_2016_CVPR,sutton2007piecewise}, that allows us to define a composite likelihood function (instead of fully normalized pdf in Eq.(\ref{eq:ll})):  
\begin{equation}
        P(\vect{Y}|\vect{X})=\prod_{q \in V}P(\vect{y}^{q}|\vect{x})\prod_{r,s \in E}P(\vect{y}^{r},\vect{y}^{s}|\vect{x}).
        \label{eq:likelihood}
\end{equation}

We include $l2$ regularization on the Unary potential. Finally, the overall cost is then given by:
\begin{equation}
        \mathop {\min}\limits_{\Omega} \lambda||\phi||_2^2-\sum_{i}^{N}
        \left[\sum_{q \in V}P_q(\vect{y}_i^{q}|\vect{x}_i)+\sum_{r,s \in E}P_{rs}(\vect{y}_i^{r},\vect{y}_i^{s}|\vect{x}_i)\right].
        \label{eq:regloglike}
\end{equation}
Where $\lambda$ defines the strength of the regularized.
However, as we show empirically, minimizing the negative log-likelihood of Eq.(\ref{eq:regloglike}) using all training data leads easily to model overfitting and, thus, poor performance. This is also due to the inherent hierarchical structure of our model (deep layers, CRF marginals and edge dependencies). 

{\bf Iterative Balanced Batch Learning.} To tackle the challenges mentioned above, we introduce an iterative balanced batch (IBB) learning approach to deal with the data imbalance during optimization of our deep structured CRF. This imbalance is highly pronounced in the number of images per training subject, average number of examples per intensity level, as well as number of different label combinations, adversely affecting the learning of  the CNN weights ($W$) and the unary ($\phi$) and pairwise ($\theta$) potential parameters, respectively. The main idea behind our IBB is to update each set of parameters with batches that are most representative of the target structure and, more importantly, balanced for that structure. To this end, when optimizing CNN weights, we generate batches ($bb_n$) that are balanced with respect to subjects in the dataset. This ensures that the learned network is not biased toward a specific subject. We adopt the same approach when creating batches for learning the marginals (balanced AU levels -- $bb_m$) and copula parameters (balanced AU co-occurrences -- $bb_e$). The learning algorithm for our network is shown in Alg.\ref{alg:bb}.
We optimize different areas of the network in each step of the algorithm. We also compute the batches in each iteration of the algorithm by sampling from the target distribution function. We apply the three step optimization iteratively, where we update the parameters of one network region and fix the remaining parameters. All updates are made with respect to the global objective defined in Eq.(\ref{eq:cost}) and we tune the validation parameter $\lambda_r$ separately for each AU. Finally, we used Stochastic GradientDescent  (SGD)  with a batch size of 128, learning rate of 0.001 and momentum of 0.9.

\begin{algorithm}
\textbf{Input:} 
        Training data: $\mathcal{D}=\{ {\vect{x}_{i},\vect{y}_{i}}\}_{i=1}^{N}$ \\
\hspace{0.95cm} Model parameters: $\Omega=\{W,\phi, \theta ,U,V \}$\\ 
\vspace{0.2cm}

\While {Eq.\ref{eq:regloglike} not converged}{
\vspace{0.2cm}
        \textbf{Step 1:} \textit{train $W$ with balanced batches:}\\
$W \leftarrow$\\
        $\mathop {\arg\!\max}\limits_{W}
        \sum_{i}^{N(bb_n)}
        P(\vect{y}_i|\vect{x}_i,W),i \in bb_n$

        \textbf{Step 2:} \textit{train $\phi$ with balanced batches:}\\
        $\forall q\in U:\phi^q \leftarrow
        \mathop {\arg\!\max}\limits_{\phi^{q}} 
        \sum_{i}^{N(bb_m)}
        \rm{Pr}(\vect{y}_i^q|\vect{x}_i,\phi^q)
        + \lambda^{q}||\phi^{q}||^2_2$
        $,i \in bb_m$

        \textbf{Step 3:} \textit{train $\theta$ with balanced batches:}\\
        $\forall (rs)\in V:\theta^{rs} \leftarrow$\\
        $\mathop {\arg\!\max}\limits_{\theta^{rs}} 
        \sum_{i}^{N(bb_e)}
        \rm{Pr}(\vect{y}_i^r,\vect{y}_i^s|\vect{x}_i,\theta^{rs})$
        $,i \in bb_e$

}
\textbf{Output:} Model parameters: $\Omega^{opt}=\{W,\phi,\theta\}$
\vspace{0.3cm}
\caption{Structured CNN Learning with balanced batches}
\label{alg:bb}
\end{algorithm}

{\bf Augmented Learning from Multiple Datasets.} As discussed in Sec.\ref{sec:intro}, leveraging data from multiple datasets efficiently is expected to further improve the AU estimation performance. To achieve this, we assume we are given $K$ datasets $\mathcal{D}\in \{D_1,D_2,\dots, D_K\}$. We then generalize the objective function of our deep structured CRFs:  
\label{sec:multi_data}
\begin{equation}
\begin{aligned}
        &P(\vect{Y}|\vect{X})=P_{D_1} \cdot P_{D_2}  \cdot \dots \cdot P_{D_K} \\
        &= \prod_{u \in \mathcal{D}}
        \prod_{q \in V_u}P(\vect{y}^{q}|x)\\
        &\times \prod_{h \in \mathcal{D}}
        \prod_{r,s \in E_h}P_h(\vect{y}^{r},\vect{y}^{s}|x).
\label{eq:cost}
\end{aligned}
\end{equation}

The key property of these sets is that they may have different AUs annotated, different dependency distributions but also contain overlapping AUs. To handle this in a principled manner, we account for the shared marginals $P(\vect{y}^{q}|x)$ -- the common  AUs,  while preserving the context-specific AU dependencies -- $P_v(\vect{y}^{r},\vect{y}^{s}|x)$ -- data-specific joints. This joint modeling is expected to result in (i) improved feature representations, and (ii) more robust learning of the (shared) marginals. To avoid bias due to the dataset order during optimization, we combined the balanced batches from all datasets, in the same manner as in the proposed IBB learning, resulting in  ${bb}_c \in \{bb_c^{1},bb_c^{2},,...,bb_c^{K}\}$, where $c\in\{n,m,e\}$. 



\subsubsection{Joint Inference}
The resulting CRF graph is an undrected graphical model that can contain loops and its potentials are not submodular. The inference of test data in this model is in general an $np$-hard problem due to the need to evaluate all possible label configurations. Because of this, we resort to approximate decoders based on the message-passing and dual decomposition algorithms. Specifically, we employed the AD3 decomposition algorithm \cite{ad3}.

\section{Experiments}
\label{sec:exp}
{\bf Datasets.} We evaluate the proposed model on two major benchmark datasets -- DISFA~\cite{mavadati2013disfa} and on the subset of the Binghamton–Pittsburgh 4D (FERA2015)~\cite{valstar2015fera}. These databases include acted and spontaneous expressions and vary in context eliciting facial expressions. The DISFA dataset contains video recordings of 27 subjects while watching YouTube videos. We performed the experiments in a subject independent setting (dividing data in training and test partition). For DISFA, we used 18 subjects for training and 9 for testing. In FERA2015 we used the official Training/Development splits. We also include the UNBC-McMaster Shoulder-Pain dataset for learning of the deep models\cite{lucey2011painful}\footnote{We use this dataset only to improve learning of our deep model; however, the evaluation results are in the supplementary materials.}. In these datasets, each frame is coded in terms of the AU intensity on a six-point ordinal scale. We use the Intra-class Correlation ICC(3,1), which is commonly used in behavioral sciences to measure agreement between annotators (in our case, the AU intensity levels). We also report the Mean Absolute Error (MAE), commonly used for ordinal prediction tasks \cite{corf,RudovicEtAlPAMI14}. \begin{figure}[tb]
\footnotesize
\setlength{\tabcolsep}{.01pt}
\begin{tabular}{p{0.1in}ccc}
\rotatebox{90}{\qquad Co-occurrence}&
\includegraphics[scale=.18]{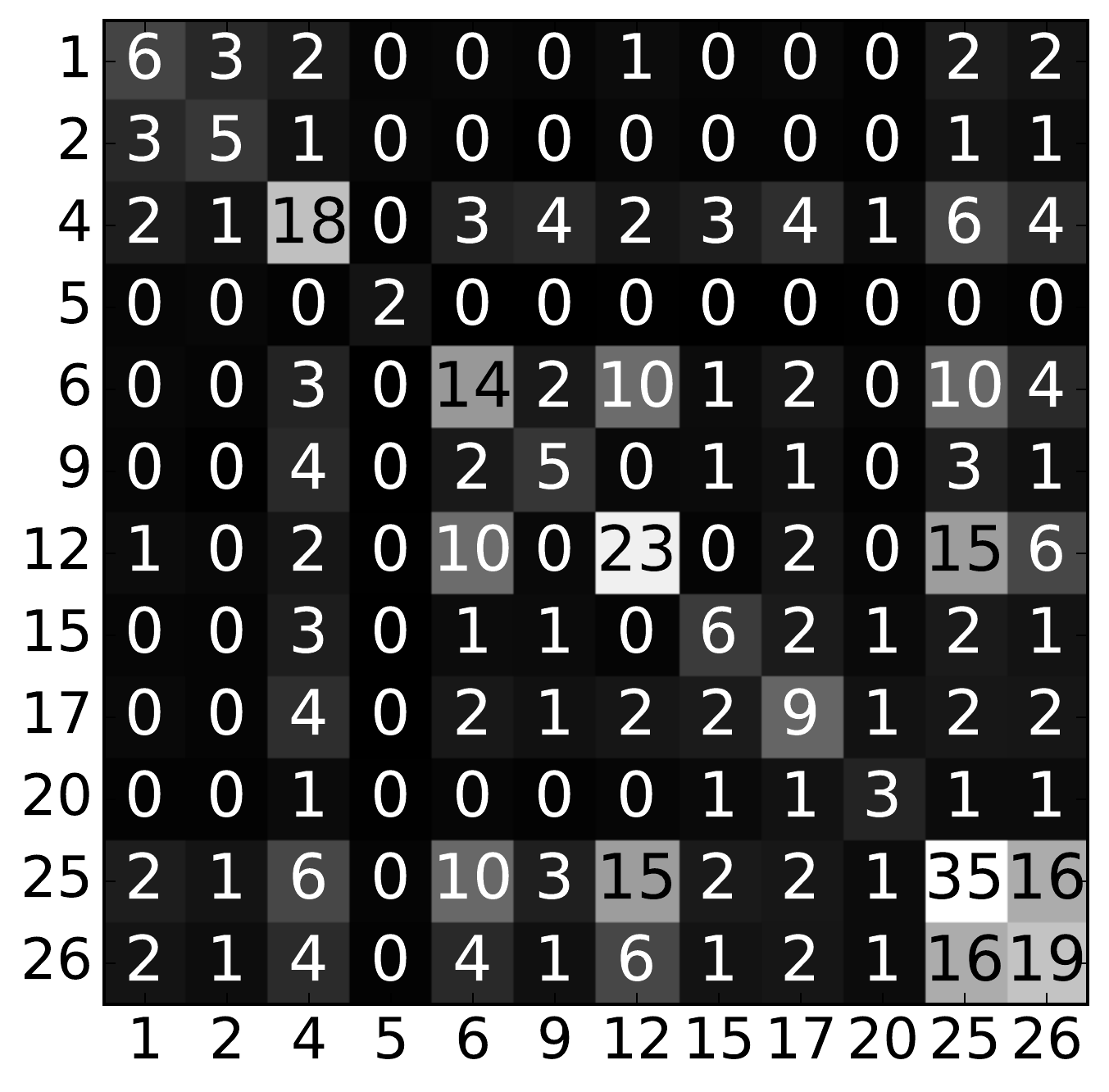} &
\includegraphics[scale=.18]{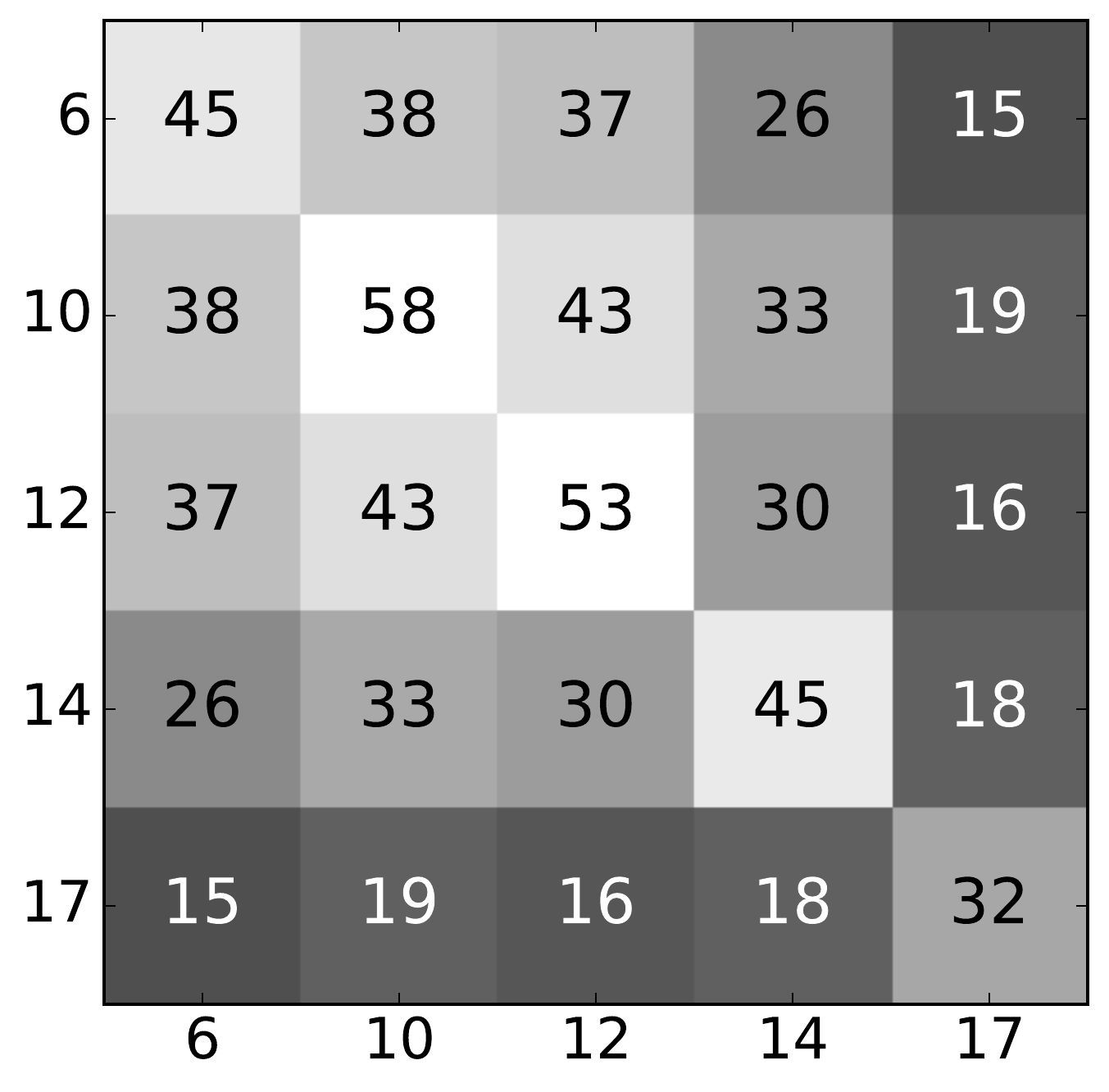} &
\includegraphics[scale=.18]{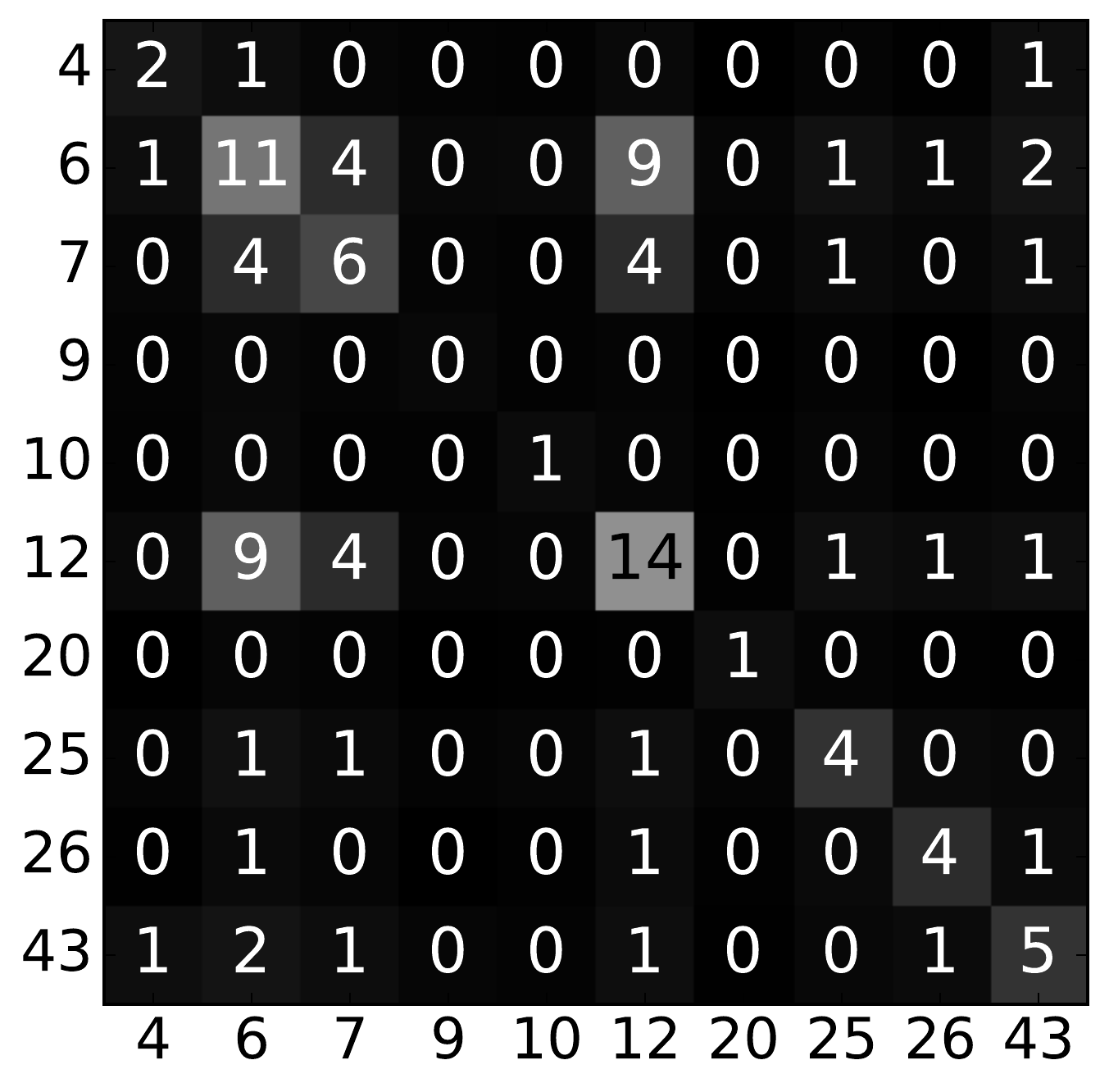} \\
\rotatebox{90}{\quad Intensity dist.}&
\includegraphics[scale=.14]{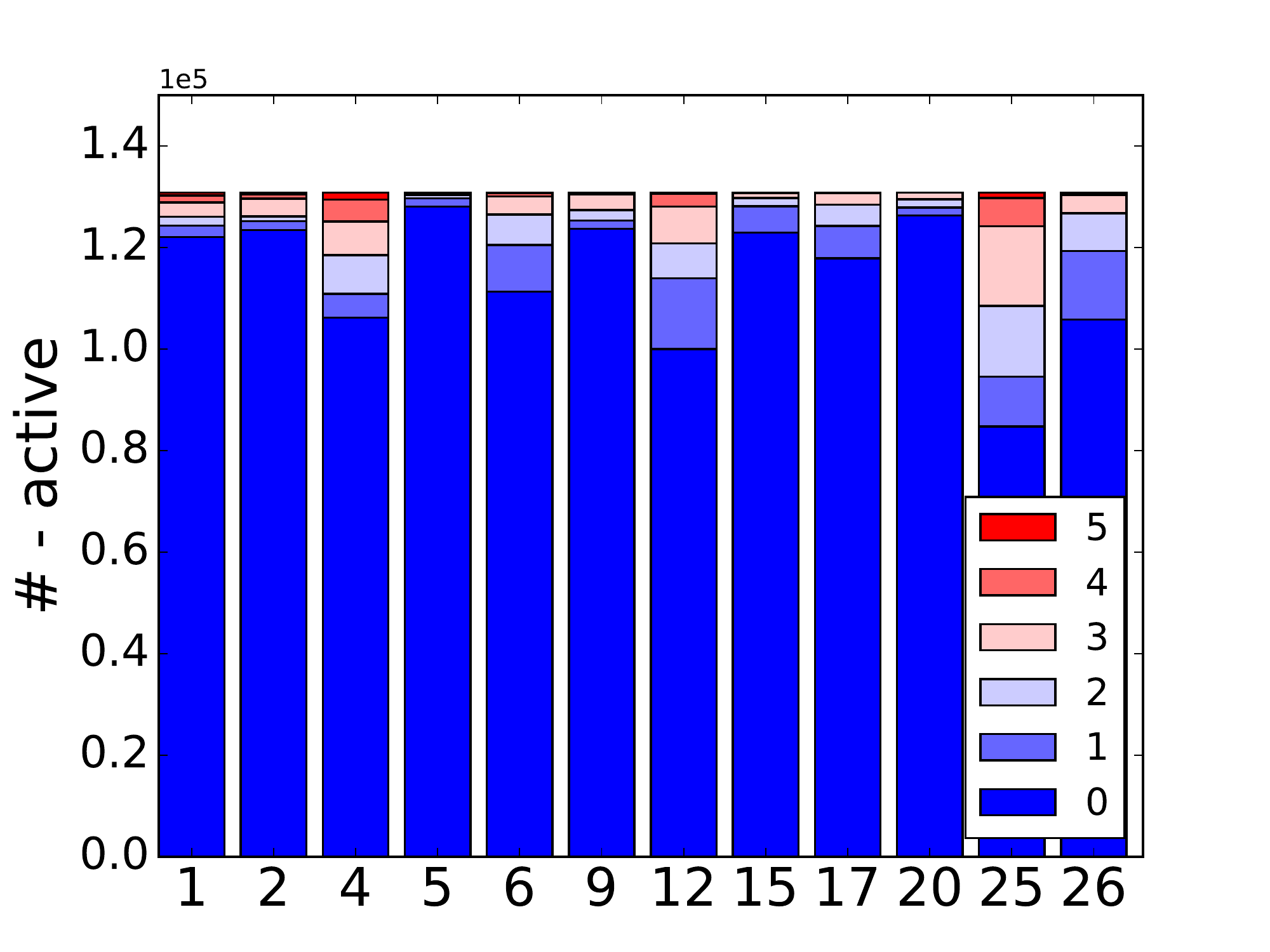} &
\includegraphics[scale=.14]{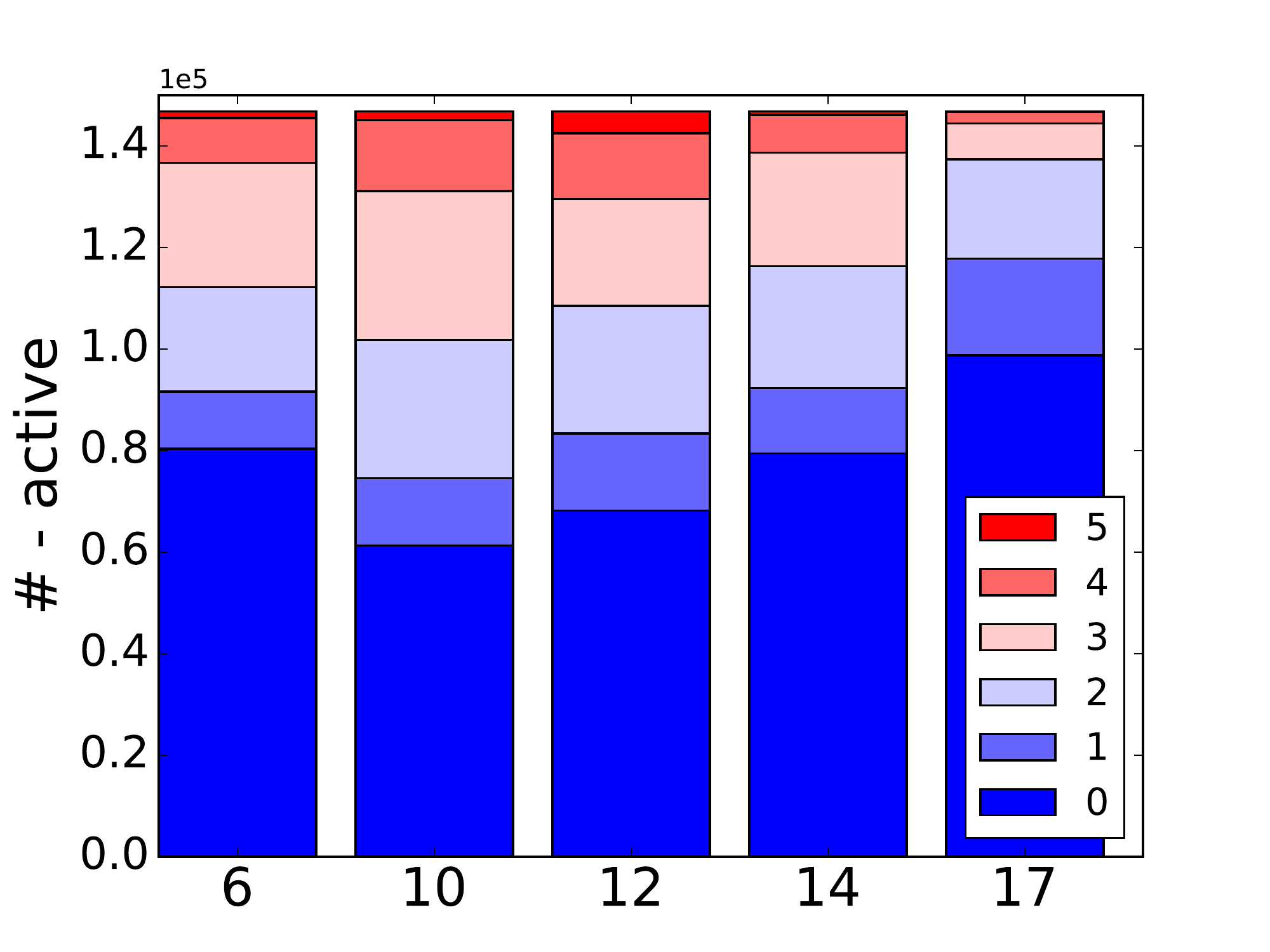} &
\includegraphics[scale=.14]{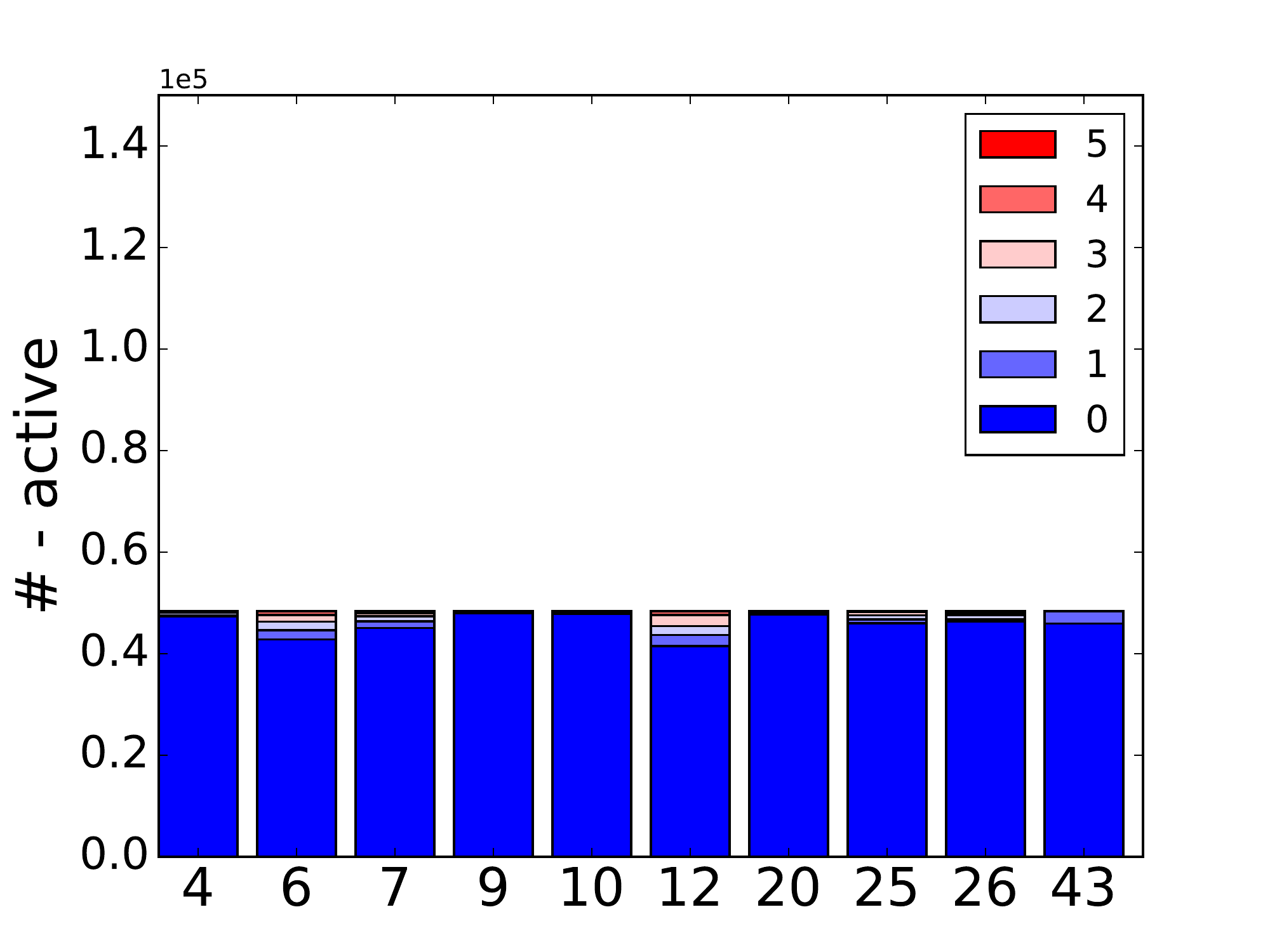} \\
\rotatebox{90}{\quad Association}&
\includegraphics[scale=.14]{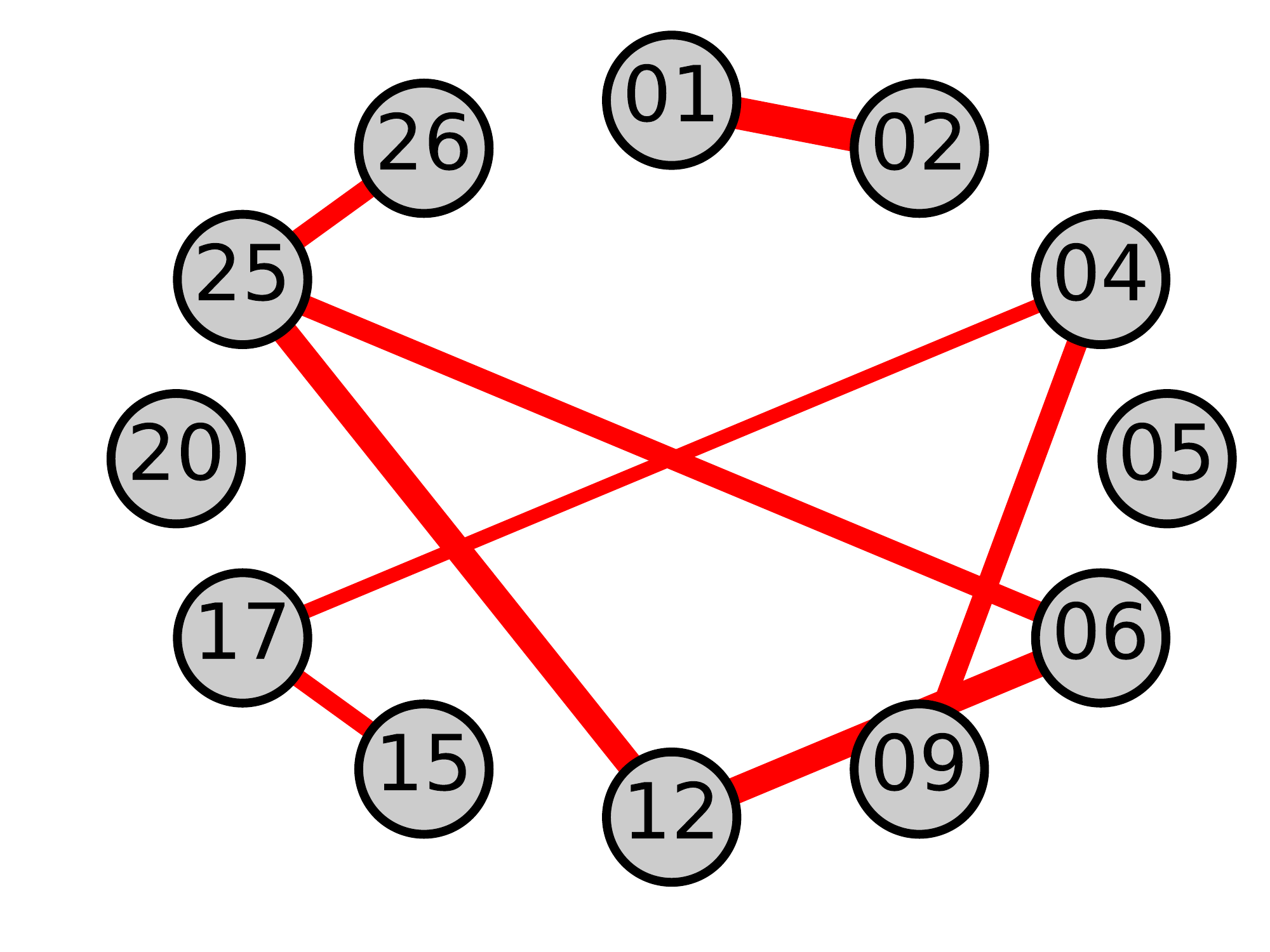} &
\includegraphics[scale=.14]{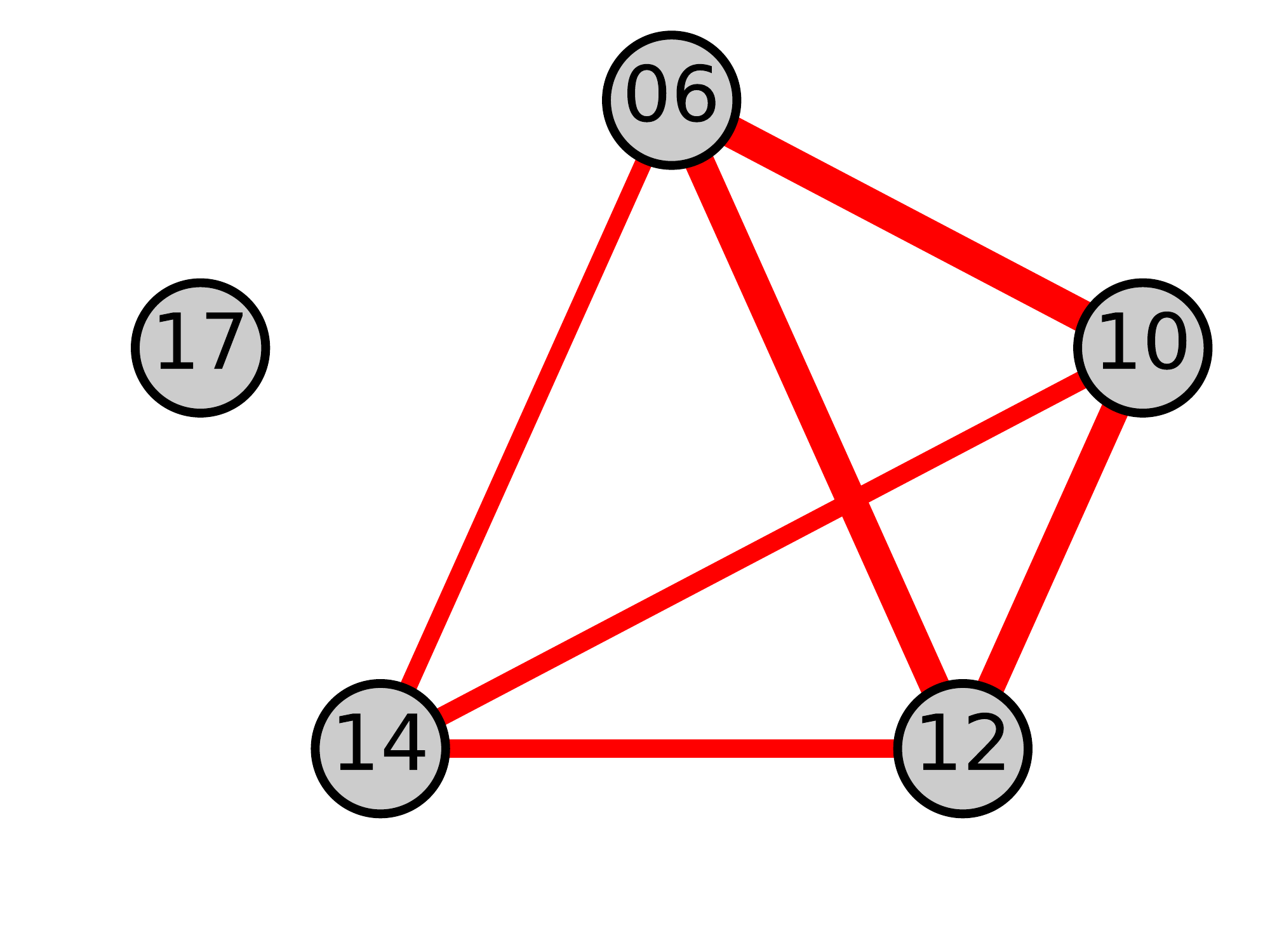} &
\includegraphics[scale=.14]{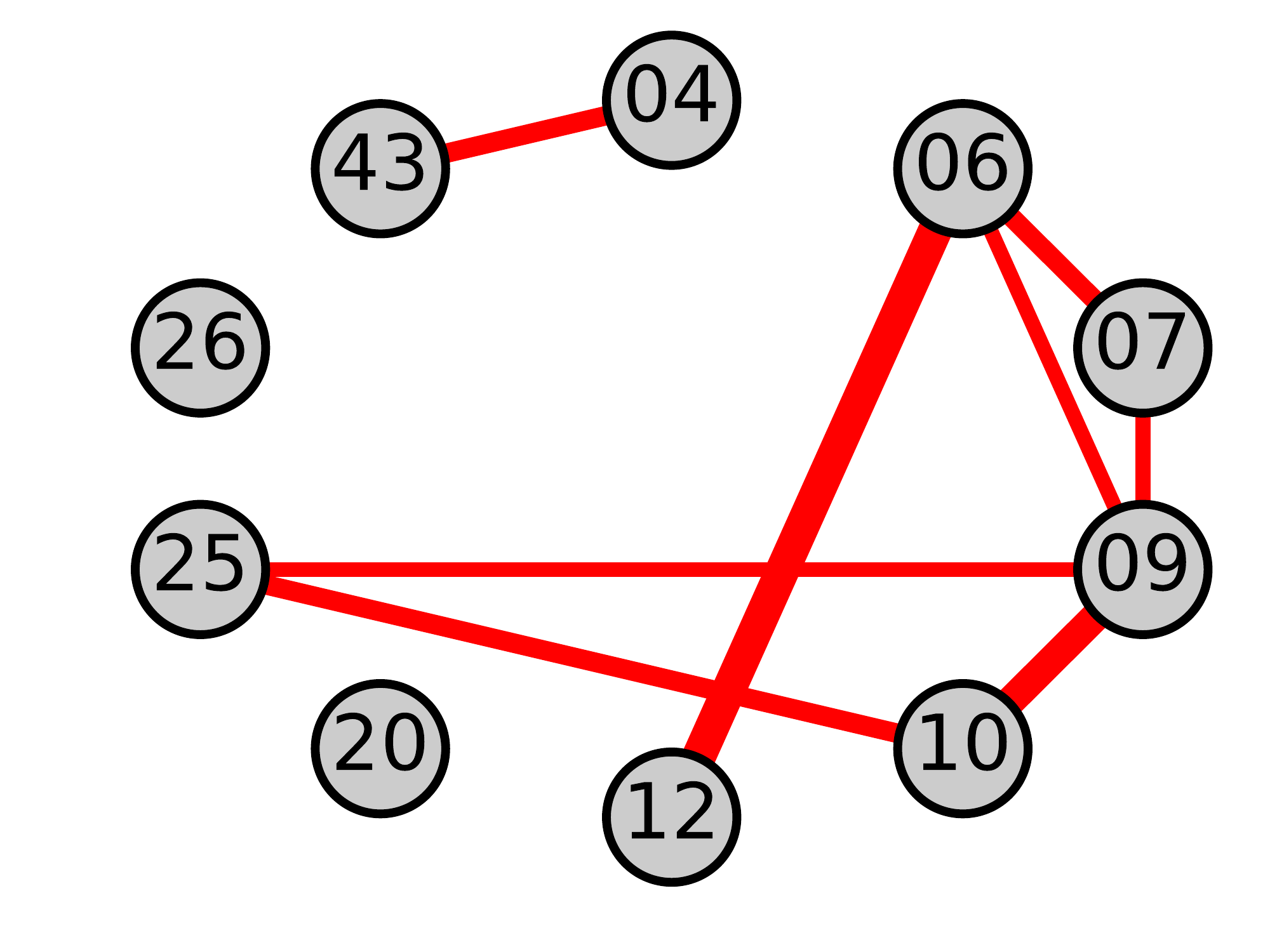} \\
&(a) DISFA & (b) FERA &(c) PAIN 
\end{tabular}
\caption{Different types of dependences of AU intensity levels in the used datasets. The thickness of the edges corresponds to the magnitude of the $\theta$ parameter. Connections with a low association ($\theta>0.05$) were removed.
}
\label{fig:graphs}
\end{figure}


{\bf Pre-Processing.} To do the basic image normalization, we used the openCV eye detector \cite{gorbenko2012face} to extract the locations of the eyes from facial images in each dataset. We then registered the 2 facial points to a reference frame (average points in each dataset) using an affine transform. We then normalized each image using per-image contrast normalization, which increases the robustness against illumination changes. Lastly, we cropped a random bounding box to 85\% of the original image size for data augmentation and robustness against displacements (as usually done in deep models).

\begin{figure}
    \begin{subfigure}{.48\linewidth}
        \centering
        \includegraphics[width=1\linewidth]{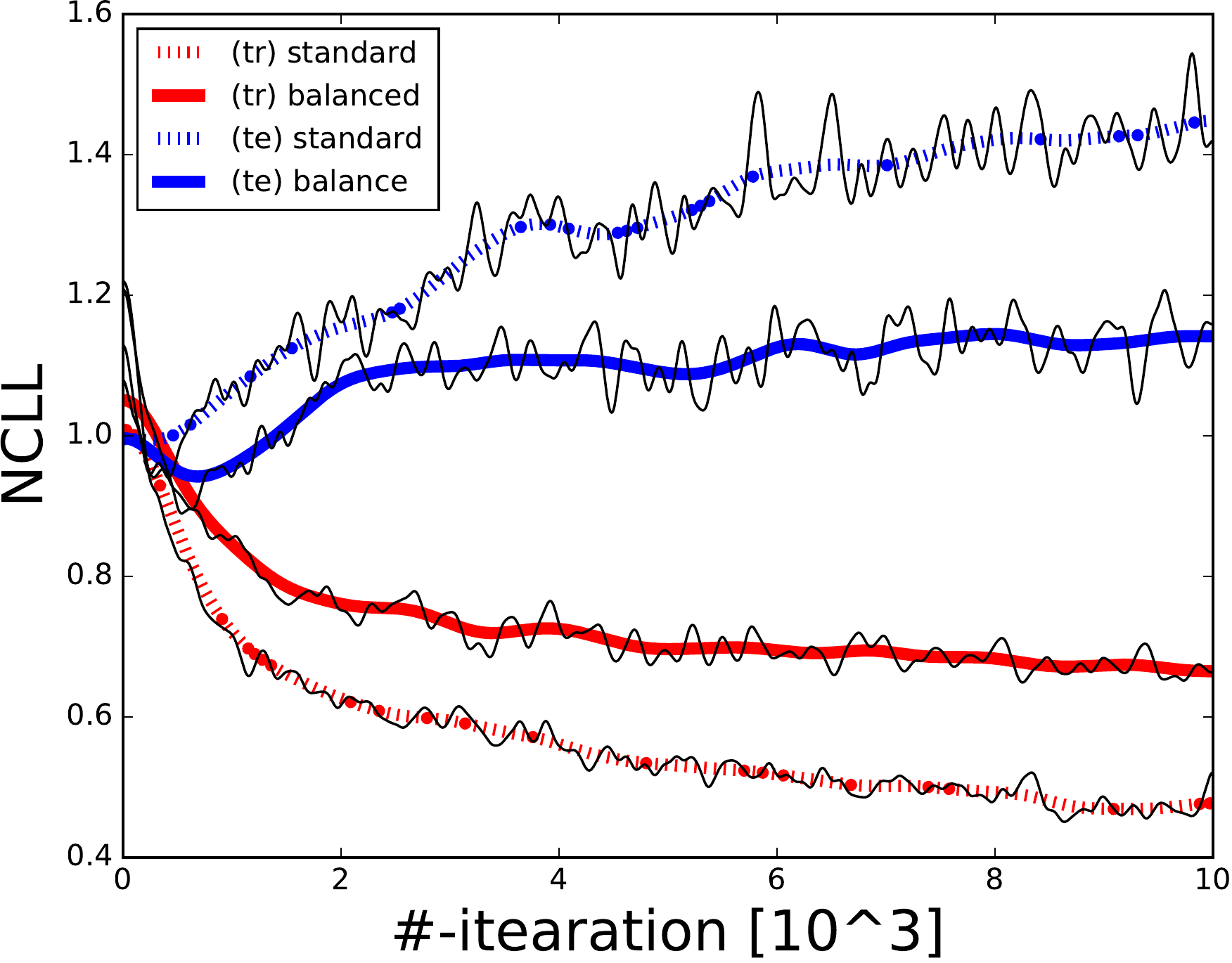}
        \caption{Learning Curves}
        \label{fig:dist_disfa}
    \end{subfigure}%
    \hfill
    \begin{subfigure}{.48\linewidth}
        \centering
        \includegraphics[width=1\linewidth]{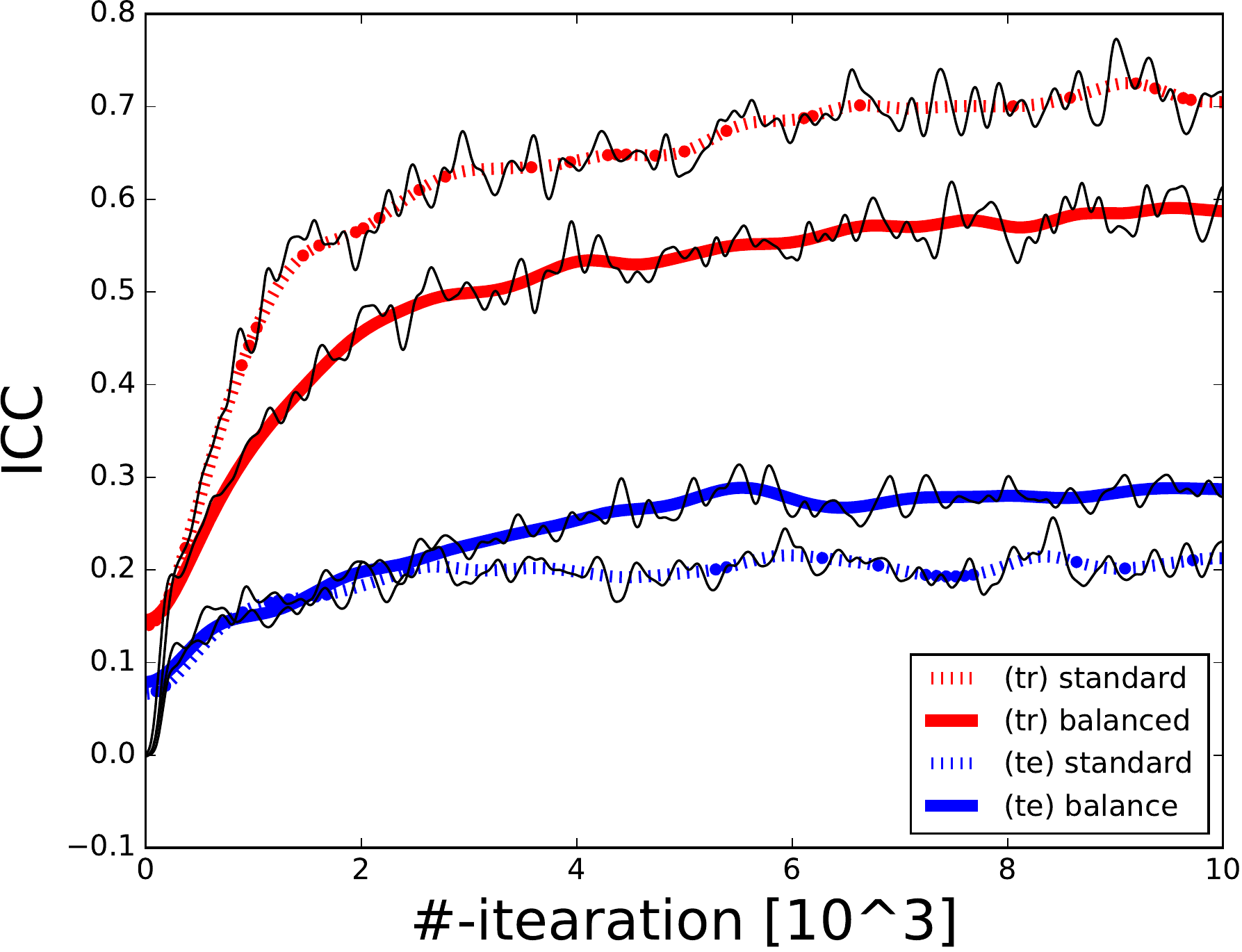}
        \caption{ICC per iteration}
        \label{fig:dist_pain}
    \end{subfigure}
        \caption{The plots show the learning curves of the \ccnn model on the DISFA dataset when balanced batches is active and inactive.} 
    \label{fig:learning_curves}
\end{figure}

{\bf Models: Baselines.} We fist conducted experiments using standard CNN architectures employed in previous works (\cite{gudi2015deep,khorrami2015deep}). The \textbf{\cnnC} model is a standard CNN with fully connected layer and softmax layer for multi-output classification. The \textbf{\ocnn} is the same network but with an ordinal classification layer. This is a special case of the proposed \ccnn where the pairwise potentials are ignored. For both methods, the weights are jointly learned and the predictions are computed independently.We conducted our experiments on a relative shallow CNN (see Fig. \ref{fig:learning}). We used 3 CNN-Layer for all our experiments but we performed cross validation to find the optimal filter size and number of channels per layer (see supplementary). Finally, the network parameters were optimized until the cost converged (see Fig.\ref{fig:learning_curves}). 

{\bf Models: CNNs.} The \textbf{\rcvprC} was introduced for AU detection tasks. It combines a basic \cnn with a customized conditional layer that has region specific weights. This feature makes the model flexible and robust by allowing the weights to be different for background and face regions, for example. The \textbf{\ocvprC} is another ordinal CNN. This network was introduced for the task of age estimation from images but can be readily applied to AU intensity estimation. \textbf{\vggC} is a widely used very deep CNN for object detection. In order to adapt it for out task, we used the pre-trained model and fine-tuned the last 3 layers for the task of AU intensity estimation. \textbf{\scnnC} is a structured CNN introduced for object detection. The linear pairwise potentials build a fully connected CRF, which is trained using piecewise \cite{sutton2007piecewise} optimization. Since this model only performs binary detection, we extended it to multi-class classification by replacing its unary potentials with the ordinal classifier. Finally, the \textbf{\ccnn} is the proposed copula conditional neural network where the pairwise potentials are defined by the copula function. The proposed model with the stepwise balanced batch optimization is named \textbf{\ccnnit}, and with the data augmentation \textbf{\ccnnit} (*). Likewise, the compared model is \textbf{\scnnit}.  

\subsection{Results}
\begin{table*}[htbp]
\scriptsize
\label{tab:results}
\caption{The intensity estimation results on the DISFA \& FERA2015 datasets for different AUs. The best results are shown in bold and in brackets. The second best results are highlighted bold. We also highlight the methods where data augmentation with multiple datasets was used with a asterix.}
\captionsetup{font=scriptsize}
\centering
\begin{tabularx}{1.0\linewidth}{|l|l|XXXXX|X|XXXXXXXXXXXX|X|}
    \hline
    \rowcolor{Gray}
    & \multicolumn{7}{c|}{FERA2015} & \multicolumn{13}{c|}{DISFA} \\
    \hline
    \rowcolor{Gray}
 &        AU:                  & 6                     & 10              & 12                     & 14               & 17              & avr.            & 1              & 2              & 4              & 5              & 6              & 9              & 12             & 15             & 17             & 20             & 25             & 26             & avr.           \\
          \hline                                                                                                                                                                                                                                                                                                                                           
    \parbox[t]{1mm}{\multirow{10}{*}{\rotatebox[origin=c]{90}{ICC(3,1)}}}
 &        \ccnnit(*)           & \textbf{[.75]}        & \textbf{.69}    & \textbf{[.86]}         & \textbf{[.40]}   & \textbf{[.45]}  & \textbf{[.63]}  & .18            & \textbf{[.15]} & \textbf{[.61]} & .07            & \textbf{[.65]} & \textbf{[.55]} & \textbf{[.82]} & \textbf{[.44]} & .37            & \textbf{[.28]} & \textbf{[.77]} & \textbf{[.54]} & \textbf{[.45]} \\
 &        \scnnC(*)            & .75                   & .67             & \textbf{[.86]}         & .39              & .42             & .62             & .16            & .12            & .43            & .06            & \textbf{.62}   & \textbf{.54}   & \textbf{[.82]} & \textbf{.43}   & .37            & \textbf{[.28]} & \textbf{[.77]} & \textbf{.53}   & \textbf{.43}   \\
 &        \ccnnit              & \textbf{[.75]}        & \textbf{.69}    & \textbf{[.86]}         & \textbf{[.40]}   & \textbf{[.45]}  & \textbf{[.63]}  & \textbf{[.20]} & .12            & \textbf{.46}   & \textbf{[.08]} & .48            & .44            & .73            & .29            & \textbf{[.45]} & .21            & .60            & .46            & .38            \\
 &        \ocnnit              & \textbf{[.75]}        & .68             & \textbf{[.86]}         & .40              & .44             & .62             & \textbf{[.20]} & .07            & \textbf{.46}   & .08            & .48            & .41            & .73            & .29            & .41            & .21            & .60            & .44            & .36            \\
 &        \ccnn                & .74                   & .67             & .85                    & \textbf{[.40]}   & \textbf{[.45]}  & .62             & .14            & .12            & .37            & \textbf{[.08]} & .46            & .44            & .64            & .25            & .37            & .09            & .58            & .31            & .32            \\
 &        \ocnn                & .73                   & .63             & .81                    & \textbf{[.40]}   & .43             & .60             & .04            & .05            & .41            & .01            & .35            & .19            & .72            & .23            & \textbf{[.45]} & .06            & .53            & .44            & .29            \\
 &        \cnnC                & .67                   & \textbf{[.69]}  & .77                    & .35              & .33             & .56             & .05            & .04            & .36            & .02            & .44            & .27            & .67            & .25            & .08            & .03            & .46            & .22            & .23            \\
 &        \rcvprC              & .62                   & .64             & .74                    & .31              & .32             & .52             & .05            & .06            & .32            & .02            & .36            & .39            & .77            & .29            & .19            & .04            & .65            & .35            & .29            \\
 &        \vggC                & .63                   & .61             & .73                    & .25              & .31             & .51             & .19            & \textbf{.14}   & .19            & .02            & .39            & .33            & .68            & .14            & .27            & .03            & .59            & .38            & .28            \\
 &        \ocvprC              & .60                   & .61             & .59                    & .25              & .31             & .47             & .03            & .07            & .01            & .00            & .29            & .08            & .67            & .13            & .27            & .00            & .59            & .33            & .20            \\
          \hline                                                                                                                                                                                                                                                                                                                                           
    \parbox[t]{1mm}{\multirow{10}{*}{\rotatebox[origin=c]{90}{MAE}}}
 &        \ccnnit(*)           & \textbf{[1.14]}       & 1.30            & .99                    & 1.65             & \textbf{[1.08]} & \textbf{[1.23]} & .87            & .63            & \textbf{[.86]} & .26            & .73            & .57            & \textbf{[.55]} & \textbf{[.38]} & .57            & .45            & \textbf{[.81]} & \textbf{[.64]} & \textbf{[.61]} \\
 &        \scnnC(*)            & 1.17                  & 1.30            & \textbf{[.97]}         & \textbf{1.60}    & 1.18            & \textbf{1.25}   & .93            & .84            & 1.05           & \textbf{.17}   & \textbf{[.71]} & .52            & .59            & \textbf{.39}   & .51            & .45            & \textbf{[.81]} & .71            & \textbf{.64}   \\
 &        \ccnnit              & 1.17                  & 1.43            & \textbf{[.97]}         & 1.65             & \textbf{[1.08]} & 1.26            & .73            & .72            & \textbf{1.03}  & .21            & .72            & \textbf{[.51]} & .72            & .43            & .50            & \textbf{[.44]} & 1.16           & .79            & .66            \\
 &        \ocnnit              & \textbf{1.15}         & \textbf{1.28}   & 1.05                   & 1.62             & 1.19            & 1.26            & .73            & \textbf{.55}   & \textbf{1.03}  & .34            & .72            & .60            & .72            & .43            & \textbf{[.47]} & \textbf{.45}   & 1.16           & .70            & .66            \\
 &        \ccnn                & 1.17                  & 1.43            & \textbf{[.97]}         & 1.65             & \textbf{[1.08]} & 1.26            & \textbf{.69}   & .72            & 1.19           & .21            & \textbf{.72}   & \textbf{[.51]} & .74            & .44            & \textbf{.48}   & .47            & 1.28           & .73            & .68            \\
 &        \ocnn                & 1.16                  & 1.32            & 1.11                   & 1.65             & 1.15            & 1.28            & 1.07           & .82            & 1.16           & .19            & .87            & .89            & .72            & .56            & .50            & .48            & 1.47           & \textbf{.69}   & .79            \\
 &        \cnnC                & 1.30                  & 1.35            & 1.28                   & 1.80             & 1.14            & 1.37            & 1.62           & 1.09           & 1.44           & .23            & .86            & .71            & .83            & .50            & .63            & .47            & 1.71           & .84            & .91            \\
 &        \rcvprC              & 1.37                  & \textbf{[1.25]} & 1.13                   & \textbf{[1.59]}  & 1.16            & 1.30            & .85            & .70            & 1.07           & .20            & .75            & .58            & \textbf{.59}   & .47            & .57            & .48            & 1.36           & .77            & .70            \\
 &        \vggC                & 1.24                  & 1.39            & 1.14                   & 1.80             & 1.19            & 1.35            & \textbf{[.68]} & \textbf{[.52]} & 1.31           & \textbf{[.16]} & .76            & .59            & .67            & .43            & .59            & .47            & 1.33           & .76            & .69            \\
 &        \ocvprC              & 1.37                  & 1.39            & 1.37                   & 1.80             & 1.19            & 1.42            & 1.05           & .87            & 1.47           & .17            & .79            & .70            & .69            & .44            & .59            & .50            & 1.33           & .86            & .79            \\
    \hline
    \end{tabularx}%
  \label{tab:results_full}%
\end{table*}%

{\bf Ordinal vs. Softmax Unary Potentials.} Table \ref{tab:results_full} shows the comparative results for the different models evaluated. On average, ordinal models largely outperform the output softmax models including related CNNs, across both measures on most of the AUs. This is particularly evident in the ICC scores, where the average difference is 7\% on the DISFA database, and 3\% on the FERA2015. We attribute this to modeling not just the different classes of intensity but also their ordinal relationship. {\bf Independent vs. Structured CNNs.} Both, \ocnn and \rcvpr achieve an ICC of 29\% on the DISFA dataset, which is the highest performance among the independent models. The \ccnn model is equivalent to \ocnn but with additional copula output for structured prediction. This results in average improvement of 4\%. This is in particular visible on AU1\&2, which are strongly correlated (see dependencies in Fig. \ref{fig:graphs}). This correlation is modeled through the copula functions with high associations parameter. {\bf The comparison with Related Deep Models.} \ocvprC performs poorly in our experiments. This model learns one binary classifier for each AU intensity level, resulting in a large number of parameters and overfitting. We achieved better results with the standard \vggC network. However, also this model does not reach comparative results with our proposed model as it does not account for ordinal intensity levels. The same applies for the \rcvprC. While both models have a significant improvement over the standard \cnnC, they fail to accurately predict ordinal intensities.

{\bf Effect of batch balancing.}
Next, we observe that the average performance of the \ccnnit model is another $3\%$ higher than that achieved by the \ccnn model (directly optimized without the data balancing). The IBB learning that we applied in \ccnnit yields a better performance on the DISFA dataset. Note that the highest improvement is made on DISFA and, in particular, on those AUs that occur infrequent (AU 1, 5, 17 and 20). As expected, we could not make this observation on the FERA database, since the labels there are relatively balanced. 

{\bf Data Augmentation.}
Finally, we analyse the contribution of additional databases for training. To study this for FERA, we augment the training data with data from DISFA and PAIN.  Similarly, for DISFA we augment the training data with FERA and PAIN.  Results obtained in this setting are indicated with (*) in Tab.~\ref{tab:results_full}.  With augmentation, the ICC for \ccnnit increases by $7\%$ on DISFA.  The largest improvement is made on AU6 and AU12, as these AUs are shared among all datasets. There is also a significant improvement on some AUs shared with only one other database (e.g., AU 4, 9, 20, and 25 that are part of PAIN data but not FERA). Lastly, we noticed a strong increase of the ICC for AU15, present only in DISFA, and a decrease on AU17 common to DISFA and FERA. This behavior is somewhat counter-intuitive but could potentially be explained by different contexts of AU17 in the two datasets (dependent in DISFA, independent in FERA). Improvements in AU15 may be the result of refined but shared feature representation. To compare with \scnnC, we use our implementation in which we minimize the same objective as in the \ccnnit model but with linear pairwise potentials instead of the copula functions. By also applying our definition of the loss for multiple datasets, we make it possible to jointly train this model on multiple datasets and compare it with the \ccnnit. We can see that, using copulas gives an improvement especially on the strongly correlated AU pairs (AU1-AU2 and AU6-AU12). This is expected, since the purpose of the Frank copula is to model pairwise correlations. On average, we obtain the highest performance with the \ccnnit model where an ICC of $0.45\%$ was reached. 

\section{Conclusion}

We  proposed a novel  Copula  CNN  deep  learning  approach  for  joint estimation of facial intensity for multiple, dependent AUs.  Specifically, we show that the ``end-to-end'' pipeline, coupled with key components for robust dependency modeling (copulas), balanced training (balanced-batch iterative), and data augmentation (across cross-context datasets) improves the performance achieved by existing structured deep models and models for estimation of facial expression intensity that fail to model non-linear dependencies in the output and also ordinal relations in the intensity levels.

\section*{Acknowledgements}
This work has been funded by the European Community Horizon 2020
[H2020/2014-2020] under grant agreement no. 645094 (SEWA). and no. 688835 (DE-ENIGMA). The work of O. Rudovic is funded by
European Union H2020, Marie Curie Action - Individual Fellowship
(EngageMe 701236), and was funded earlier by the European Community Horizon 2020 [H2020/2014-2020] under grant agreement no. 688835 (DE-ENIGMA). The work of Vladimir Pavlovic is funded by the National Science Foundation under Grant no. IIS0916812.


{\small
\bibliographystyle{ieee}
\bibliography{bibfile}
}

\end{document}